\documentclass{article} 
\usepackage[final]{colm2026_conference}
\usepackage{amsmath}
\usepackage{amssymb}
\usepackage{microtype}
\usepackage{hyperref}
\usepackage{url}
\usepackage{booktabs}
\usepackage{graphicx}
\usepackage{xcolor}
\usepackage{tabularx}
\usepackage{multirow}
\usepackage{subcaption}
\usepackage[most]{tcolorbox}
\usepackage{listings}

\DeclareFontFamily{OMS}{zi4}{}
\DeclareFontShape{OMS}{zi4}{m}{n}{<->ssub * cmsy/m/n}{}

\definecolor{csviolet}{HTML}{645396}
\definecolor{cslilac}{HTML}{ACA0C2}
\definecolor{csgold}{HTML}{B9975B}
\definecolor{cspeach}{HTML}{FFC19B}

\usepackage{lineno}

\definecolor{darkblue}{rgb}{0, 0, 0.5}
\hypersetup{colorlinks=true, citecolor=darkblue, linkcolor=darkblue, urlcolor=darkblue}

\title{FinHardBench: Can LLMs Generate Latency-Aware Hardware for Financial Computing?}

\author{Weimin Fu$^{1}$ \quad Hejia Zhang$^{2}$ \quad Minghao Shao$^{3,4}$ \quad Zeng Wang$^{3}$ \quad Johann Knechtel$^{4}$ \\
\textbf{Ozgur Sinanoglu$^{4}$ \quad Muhammad Shafique$^{4}$ \quad Ramesh Karri$^{3}$ \quad Xiaolong Guo$^{1}$} \\
$^1$Lehigh University \quad $^2$University of California, San Diego \\
$^3$NYU Tandon School of Engineering\quad $^4$NYU Abu Dhabi \\
\texttt{\{wef326,xig426\}@lehigh.edu},\quad \texttt{hez024@ucsd.edu}\\\texttt{\{{shao.minghao, zw3464, johann, ozgursin, muhammad.shafique, rkarri\}@nyu.edu}}}

\begin{document}

\ifcolmsubmission
\linenumbers
\fi

\maketitle

\begin{abstract}
Can large language models generate not just correct, but fast hardware? This paper investigates the question in financial FPGA design, where 5--10 nanoseconds of latency determines competitive advantage and designs iterate continuously as protocols, strategies, and regulations evolve. FinHardBench, a benchmark of 33 financial computing tasks, is presented together with three experiments that mirror the real-world FPGA iteration cycle: generating new modules from specifications, tuning system-level configurations across a 6-stage trading pipeline, and adapting existing modules to specification changes. Evaluation of six LLMs on 1530+ experiment rounds yields three findings: (1) models achieve 19--61\% functional correctness with timing degradation up to 13.7$\times$ on specific tasks; (2) in system-level design space exploration, top LLMs converge to the optimal configuration with higher reliability than random search, simulated annealing, and Bayesian optimization baselines (5/5 seeds vs.\ 0--4/5 at the same 24-round budget); (3) strategy-level specification changes remain unsolved for most models. Across the six models, generation and DSE rankings overlap moderately: the strongest code generator is not the fastest architecture optimizer, and the weakest code generator (MiniMax M2.7) still reaches the system optimum on 4 of 5 seeds. On the tasks in FinHardBench, difficulty tracks training data pattern availability more closely than abstraction level. FinHardBench is released as an open-source benchmark.
\end{abstract}

\section{Introduction}
\label{sec:intro}

A trading firm updates its FPGA-based order execution system. The exchange has modified its market data protocol, the strategy team has replaced a moving-average indicator with a MACD signal, and the risk department has mandated a new circuit breaker. Each change requires modifying RTL code, re-synthesizing, verifying timing closure, and redeploying. This cycle repeats on weekly to monthly cadence. Manual iteration takes weeks; the strategy team needs results in hours.

This scenario motivates a specific question: can LLMs accelerate the financial FPGA iteration cycle? The answer depends on three distinct capabilities. \textbf{Module generation}: given a specification for a new component, can an LLM produce functionally correct, timing-efficient Verilog? \textbf{System configuration}: given a multi-stage trading pipeline with per-stage knobs, can an LLM select configurations that minimize end-to-end latency? \textbf{Specification adaptation}: given an existing module and a changed requirement, can an LLM correctly modify the code while preserving timing quality?

Prior work has studied the first capability in isolation. VerilogEval~\citep{liu2023verilogeval}, RTLLM~\citep{lu2024rtllm}, and ResBench~\citep{guo2025resbench} evaluate LLM-generated Verilog on functional correctness and resource utilization. While recent concurrent work such as CVDP~\citep{pinckney2025cvdp} includes isolated timing-related tasks, no existing benchmark systematically evaluates post-P\&R timing quality across a domain-specific task suite, and none centers on financial computing, a domain where FPGA is the dominant platform~\citep{boutros2017hft,lockwood2012fpga}, with nanosecond-level latency requirements (FAST decoding: 36--56\,ns~\citep{kan2021hls}; order book: 26--209\,ns~\citep{liu2024orderbook}). The second and third capabilities have not been systematically benchmarked. A parallel line of work applies LLMs to alpha factor mining---the software-level discovery of predictive trading signals~\citep{shi2025alphaforge,tang2025alphaagent,luo2026alphabench}; FinHardBench addresses the downstream problem of implementing such computations as latency-aware hardware.

This paper presents FinHardBench, a benchmark of 33 financial FPGA tasks across five abstraction layers, together with three experiments that mirror the iteration cycle described above. The benchmark evaluates not only functional correctness but also post-P\&R timing on Lattice ECP5 FPGAs, measuring critical-path delay between domain-specific I/O pins at 5--10\,ns granularity.

The evaluation of six LLMs across 1530+ experiment rounds produces a central finding: \textbf{the three capabilities overlap only moderately across models}. MiniMax M2.7 generates the worst code (19\% functional pass) but converges to near-optimal system configurations. Claude Sonnet 4.6 generates the best code (61\% pass), discovers the optimal system configuration fastest (Round~11 on average), and is the only model in this evaluation that achieves a nonzero strategy-adaptation pass rate (42\% on strategy changes). No tested model succeeded at the pipeline-insertion timing-optimization task. Within 24 rounds, every model reaches the 107.5\,ns optimum in at least 1 of 5 seeds, but convergence speed varies by 2$\times$ (Claude: Round~11, Mistral: Round~23).

\begin{figure}[!b]
    \centering
    \includegraphics[width=0.93\textwidth]{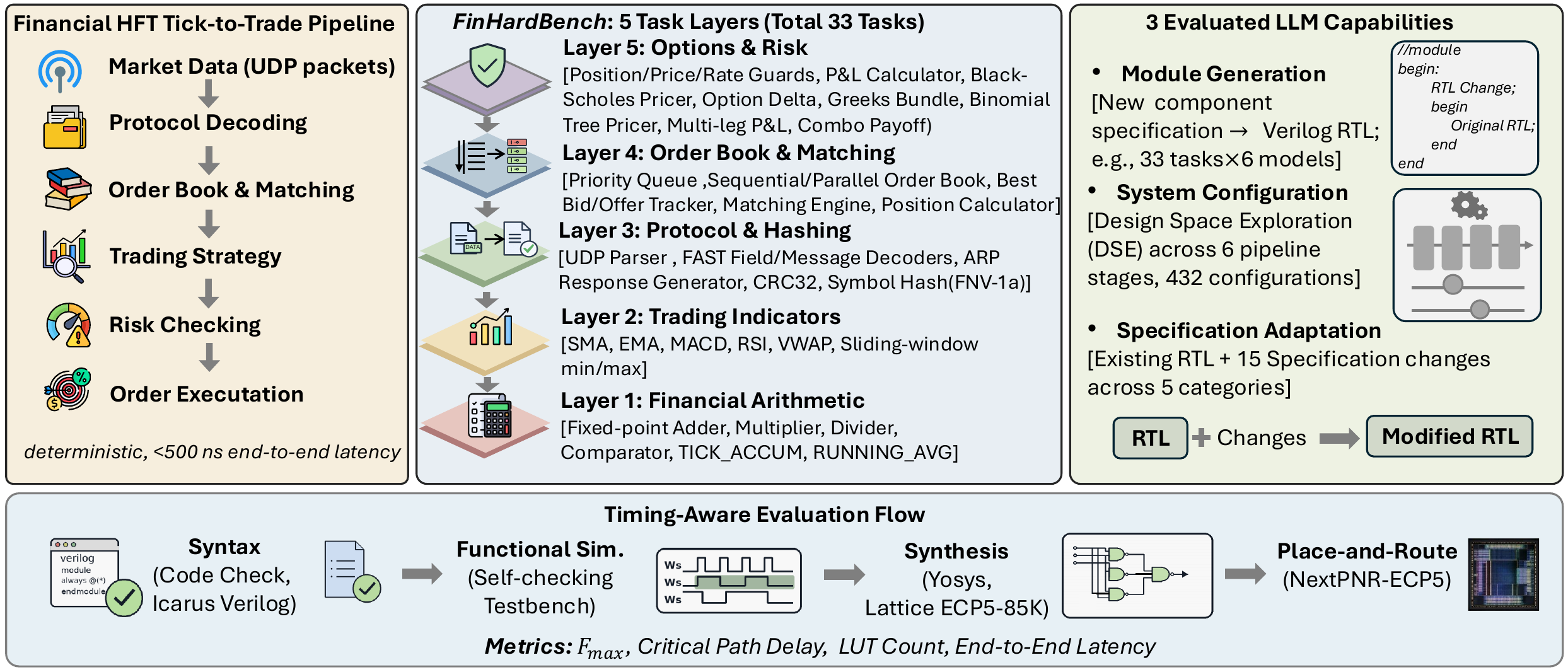}
    \caption{\textbf{Overview of FinHardBench.}
    A financial HFT tick-to-trade pipeline grounds five task layers (33 tasks) evaluated along three LLM capabilities---module generation, system configuration, and specification adaptation---through a timing-aware flow (syntax, simulation, synthesis, P\&R) using $F_{\max}$, critical path delay, LUT count, and end-to-end latency as metrics.}
    \label{fig:overview}
\end{figure}

LLM-assisted hardware design is not a single problem to be solved by a single model. FinHardBench provides the infrastructure to measure progress on each dimension separately. Figure~\ref{fig:overview} provides an overview.

\section{Background}
\label{sec:background}

\textbf{FPGA in financial computing.}
High-frequency trading (HFT) systems process market data and execute orders within microseconds. The tick-to-trade pipeline follows a fixed topology: network packet reception, market data protocol decoding, order book reconstruction, trading signal computation, risk checking, and order execution~\citep{yoo2023lighttrader}. FPGAs serve as the dominant platform, providing deterministic, sub-microsecond latency that general-purpose processors cannot match~\citep{boutros2017hft}, with 270--5400$\times$ speedup over CPUs on option pricing workloads~\citep{mahony2024fpga}. Production systems on Xilinx Alveo/UltraScale+ achieve end-to-end latencies below 500\,ns~\citep{alveo2019}. Financial exchanges transmit market data using specialized binary protocols including FAST~\citep{kan2021hls}, ITCH, SBE, and XDP~\citep{lockwood2012fpga}. Three factors drive continuous FPGA redesign: \textbf{protocol evolution}, \textbf{strategy adaptation}, and \textbf{hardware migration}. Manual RTL iteration for each change takes days to weeks; LLMs offer a potential path to compress this cycle.

\textbf{FPGA design flow and timing closure.}
The FPGA flow consists of \emph{synthesis}, \emph{place-and-route} (P\&R), and \emph{bitstream generation}; a design that passes functional simulation may still fail timing closure due to long combinational paths or routing congestion. Two metrics quantify timing: \textbf{Fmax}, set by the longest register-to-register path, and \textbf{critical path delay}, the logic-plus-routing delay between two specific pins---for financial FPGA, a domain-specific pair such as \texttt{order\_valid}$\to$\texttt{match\_valid} in a matching engine. At the 5--10\,ns scale of HFT, a 2\,ns difference is competitively meaningful. In a multi-stage pipeline, system Fmax is the minimum per-stage Fmax; end-to-end latency is total pipeline depth divided by system Fmax.

\textbf{LLMs for HDL generation.}
LLMs generate Verilog from natural language zero-shot (GPT, Claude, Gemini), after fine-tuning on hardware code (VeriGen~\citep{thakur2023verigen}, RTLCoder~\citep{liu2024rtlcoder}, CodeV~\citep{zhao2024codev}), or through domain adaptation on proprietary EDA corpora~\citep{liu2023chipnemo}; existing benchmarks such as VerilogEval~\citep{liu2023verilogeval} and RTLLM~\citep{lu2024rtllm} measure functional correctness via pass@$k$ (Table~\ref{tab:benchmarks}).

\begin{table}[t]
\centering
\caption{Feature comparison with existing HDL generation benchmarks. The first two axes are shared with prior work; the remaining five are where FinHardBench adds. To our knowledge, no prior benchmark covers the system-level DSE and specification-adaptation axes at this granularity.}
\label{tab:benchmarks}
\small
\setlength{\tabcolsep}{3.5pt}
\begin{tabular}{lccccccc}
\toprule
 & Spec-to- & Code qual. & Financial & Post-P\&R & Domain & System & Spec \\
 & RTL eval & (res./QoR) & domain & timing & crit.-path & DSE & adapt. \\
\midrule
VerilogEval (156) & \checkmark & -- & -- & -- & -- & -- & -- \\
RTLLM (50) & \checkmark & -- & -- & -- & -- & -- & -- \\
ResBench (56) & \checkmark & \checkmark (LUT) & partial & -- & -- & -- & -- \\
CVDP (783) & \checkmark & \checkmark (QoR) & -- & partial & -- & -- & -- \\
VeriPPA & partial & \checkmark (PPA) & -- & ASIC & -- & -- & -- \\
\midrule
\textbf{FinHardBench} & \checkmark & \checkmark & \checkmark & \checkmark & \checkmark & \checkmark & \checkmark \\
\bottomrule
\end{tabular}
\end{table}

FinHardBench comprises three layers that correspond to the iteration modes used in production financial FPGA work. The \textbf{Bench} layer provides 33 module-level tasks with post-P\&R timing evaluation (Fmax, critical path delay); it shares the spec-to-RTL pattern with prior HDL benchmarks and is included for comparability, although green-field module generation is the least frequent iteration mode in production. The \textbf{System} layer defines a 6-stage pipeline with 432 configuration combinations for system-level design space exploration (DSE), in which an LLM iteratively selects per-stage microarchitectural options to minimize end-to-end latency; this per-stage knob tuning on a stable topology is the dominant production iteration mode. The \textbf{Adapt} layer tests targeted code modification in response to 15 specification changes across 5 categories, the mode driven by protocol, indicator, and regulatory changes. To our knowledge, no prior benchmark covers the System and Adaptation layers at this granularity.

\section{FinHardBench}
\label{sec:benchmark}

\begin{figure}[t]
  \centering
  \begin{minipage}[c]{0.66\textwidth}
    \centering
    \small
    \setlength{\tabcolsep}{3pt}
    \begin{tabularx}{\linewidth}{@{}c l X@{}}
      \toprule
      \textbf{Ly.} & \textbf{Category} & \textbf{Description} \\
      \midrule
      L1 & Fin.\ Arithmetic  & Fixed-point arithmetic for pricing and execution. \\
      L2 & Trading Indicators & Technical indicators (SMA, MACD, RSI, etc.)\ from streaming ticks. \\
      L3 & Protocol \& Hash   & UDP and FAST protocol decoding; CRC and symbol hashing. \\
      L4 & Order Book         & Sorted book maintenance, BBO tracking, and order matching. \\
      L5 & Options \& Risk    & Risk guards, P\&L calculation, and option pricing (BS, binomial). \\
      \bottomrule
    \end{tabularx}
  \end{minipage}\hfill
  \begin{minipage}[c]{0.26\textwidth}
    \centering
    \definecolor{pieborder}{HTML}{5E548E}
    \fcolorbox{pieborder}{white}{\includegraphics[width=\dimexpr\linewidth-2\fboxsep-2\fboxrule]{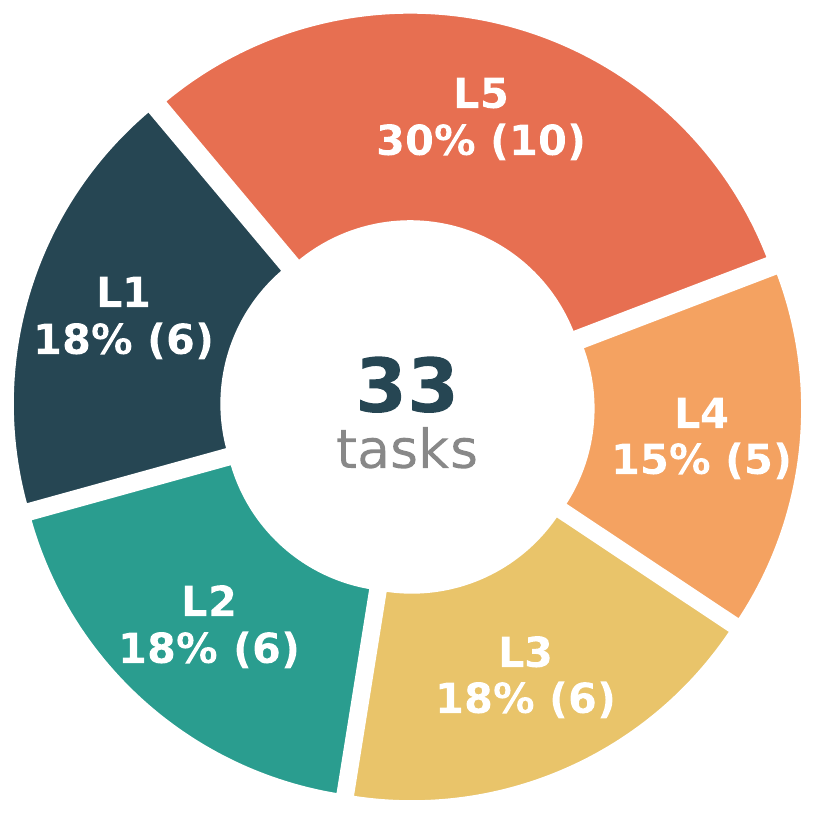}}
  \end{minipage}
  \caption{Task distribution and layer descriptions of FinHardBench.}
  \label{fig:task-distribution}
  \vspace{-4mm}
\end{figure}

\textbf{Benchmark Composition.} Figure~\ref{fig:task-distribution} summarizes the 33 tasks across five layers that mirror a production tick-to-trade pipeline, from the fixed-point arithmetic primitives underpinning all downstream computation (L1) up to Options \& Risk (L5), the largest group with 10 tasks spanning pre-trade risk checks, P\&L tracking, and analytical and numerical option pricing.

\textbf{Per-task structure.}
Two tasks illustrate the domain-specific design: \textsc{Tick\_Accum} (L1) accumulates price changes only when they cross a tick boundary, filtering sub-tick market noise; \textsc{Running\_Avg} (L1) computes TWAP with a dynamically growing divisor, requiring multi-cycle iterative division. The ten L5 option pricing and risk tasks are cataloged in Appendix~\ref{app:l5}. Each task ships with five files: a natural language specification, a self-checking testbench with pass/fail verdict, one or more golden reference implementations, \texttt{critical\_io.json} defining the domain-specific timing path (e.g., \texttt{order\_valid}$\to$\texttt{match\_valid} for the matching engine), and \texttt{knobs.json} listing parameterized design options (data width, buffer depth, architecture variant).

\textbf{Evaluation pipeline and metrics.}
Four stages process each generated design: (1) syntax check via Icarus Verilog, (2) functional simulation against the testbench, (3) synthesis via Yosys targeting Lattice ECP5-85K, (4) P\&R via NextPNR-ECP5 extracting Fmax, LUT count, and per-path-type delays. The open-source ECP5 toolchain ensures reproducibility; production HFT systems use Xilinx UltraScale+/Alveo, and a 13-design cross-toolchain pilot on Kintex UltraScale+ shows strong Fmax rank agreement with the ECP5 results ($\rho=0.982$; Appendix~\ref{app:vivado}). For designs passing simulation, timing quality is measured relative to golden references via \textbf{Fmax ratio} ($\text{Fmax}_\text{LLM} / \text{Fmax}_\text{golden}$) and \textbf{timing ratio} ($\text{CritPath}_\text{LLM} / \text{CritPath}_\text{golden}$), both measured at domain-specific critical I/O pins.

\section{Experiments}
\label{sec:experiments}

Six LLMs are evaluated via OpenRouter: GPT-5.4 (OpenAI), Claude Sonnet 4.6 (Anthropic), Gemini 3.1 Flash Lite (Google), DeepSeek V3.2 (DeepSeek), MiniMax M2.7 (MiniMax), and Mistral Large 2512 (Mistral AI). The evaluation infrastructure uses Yosys~0.57 for synthesis, NextPNR for ECP5-85K P\&R, and Icarus Verilog~12.0 for simulation, with SQLite-backed parallelism (20 workers). Three experiments map to the three stages of the financial FPGA iteration cycle. Each fixes its prompting paradigm so that observed differences reflect the model rather than the harness: Experiment~1 uses \emph{single-shot} prompting; Experiments~2 and~3 use \emph{structured feedback-driven} prompting (fixed context, trial history, no tool calls). \emph{Agentic} prompting is a third paradigm: a pilot (Appendix~\ref{app:agentic}) shows off-the-shelf coding agents passing hard tasks (VWAP, RSI) where single-shot fails, so the single-shot results below reflect a one-call budget rather than a capability ceiling.

\subsection{Experiment 1: Module generation}
Each LLM receives a task's natural language specification and produces a complete Verilog module, evaluated through all four pipeline stages. The prompt includes only the specification text; no golden reference or testbench is shown. Scale: 6 models $\times$ 33 tasks $\times$ 3 trials = 594 evaluations.

\begin{table}[t]
\centering
\caption{Module generation: pass rates across 33 tasks $\times$ 3 trials. Metrics form a consecutive-AND chain (each column is a subset of the previous one): Syn\% = syntax parse succeeds; Sim\% = Syn \emph{and} the testbench passes; Synth\% = Sim \emph{and} synthesis completes; Routed\% = Synth \emph{and} P\&R succeeds. Fmax (MHz) and LUTs averaged over all designs completing FPGA P\&R, independent of simulation outcome.}
\label{tab:exp1_summary}
\begin{tabular}{lrrrrrr}
\toprule
Model & Syn\% & Sim\% & Synth\% & Routed\% & Fmax & LUTs \\
\midrule
GPT-5.4 & 92 & \textbf{61} & 57 & 54 & 141 & 1573 \\
Claude Sonnet 4.6 & 86 & \textbf{61} & \textbf{61} & \textbf{57} & 165 & 1096 \\
Gemini 3.1 Flash Lite & \textbf{94} & 46 & 44 & 41 & 181 & 897 \\
DeepSeek V3.2 & 78 & 25 & 24 & 23 & 149 & 892 \\
Mistral Large & 78 & 23 & 23 & 20 & 149 & 704 \\
MiniMax M2.7 & 40 & 19 & 19 & 19 & 231 & 382 \\
\bottomrule
\end{tabular}
\end{table}

\begin{table}[t]
\begin{minipage}[t]{0.46\textwidth}
\centering
\caption{Critical-path timing for designs passing simulation vs.\ golden references. Ratio $>$1.0: LLM design is slower.}
\label{tab:timing_gap}
\small
\setlength{\tabcolsep}{3pt}
\resizebox{\linewidth}{!}{%
\begin{tabular}{lrrr}
\toprule
Task & Golden (ns) & LLM (ns) & Ratio \\
\midrule
UDP\_RX & 3.5 & 3.3 & 0.94$\times$ \\
EMA & 12.6 & 12.2 & 0.97$\times$ \\
SMA & 7.0 & 7.1 & 1.02$\times$ \\
RATE\_LIMIT & 3.9 & 4.0 & 1.02$\times$ \\
ORDER\_BOOK & 22.1 & 23.5 & 1.06$\times$ \\
MACD & 12.3 & 14.4 & 1.17$\times$ \\
PRIO\_Q & 9.3 & 12.7 & 1.37$\times$ \\
FP\_DIV & 5.1 & 8.1 & 1.59$\times$ \\
MINMAX & 4.2 & 10.4 & 2.45$\times$ \\
RSI & 99.0 & 358.2 & 3.62$\times$ \\
VWAP & 5.3 & 72.7 & \textbf{13.67$\times$} \\
\bottomrule
\end{tabular}}
\end{minipage}\hfill
\begin{minipage}[t]{0.51\textwidth}
\centering
\caption{Deployable-but-wrong designs on L2 trading-indicator tasks (18 attempts each). ``Inflation'' = routed-but-wrong / routed.}
\label{tab:deployable}
\small
\setlength{\tabcolsep}{3pt}
\begin{tabular}{lrrrr}
\toprule
Task & Sim $\checkmark$ & Routed & Wrong & Inflation \\
\midrule
RSI    &  1 & 10 &  9 & 90\% \\
VWAP   &  2 & 11 &  9 & 82\% \\
EMA    &  5 & 11 &  6 & 55\% \\
MACD   &  8 & 15 &  7 & 47\% \\
SMA    & 14 & 16 &  2 & 13\% \\
MINMAX & 13 & 13 &  0 &  0\% \\
\bottomrule
\end{tabular}
\end{minipage}
\end{table}

In Table~\ref{tab:exp1_summary}, the toolchain metrics are defined as a consecutive-AND chain, so Sim\%---the only column with correctness semantics---upper-bounds every later column. Under this definition, Sim\% and Routed\% are within 1--7 percentage points for every model; the residual gap is the fraction of sim-passing designs that fail to route, typically due to resource exhaustion on large designs. The converse population is far larger: 206 of 594 attempts (34.7\%) produce HDL that synthesizes and routes but fails simulation. Producing structurally valid, synthesizable HDL is easier than producing functionally correct HDL for all six models, and toolchain-level success alone is not a deployment gate; Section~\ref{sec:deployable} analyzes this population.

Pass rates vary across layers but do not follow a monotonic difficulty gradient. Risk/compliance tasks (72\%) involve threshold comparisons and guard logic with well-represented patterns in training data. Order book operations (31\%) require concurrent sorted data structure manipulation. Arithmetic tasks (43\%) suffer when efficient fixed-point division is needed (VWAP, RSI), a pattern largely absent from open-source Verilog corpora. These results suggest that the availability of the required implementation pattern in public training data affects difficulty more than abstraction level does. Several tasks achieve 0\% across all models (BBO\_TRACK, FAST\_MSG, TICK\_ACCUM, BS\_CALL\_PRICE), and the remaining option pricing tasks pass in at most 2 of 18 attempts (OPTION\_DELTA, GREEKS\_BUNDLE, BINOM\_PRICER); these require polynomial approximations of transcendental functions in fixed-point arithmetic. Appendix~\ref{app:l5} covers all ten L5 tasks and shows that the failures are algorithm substitutions against explicitly named canonical algorithms, not testbench strictness or prompt under-specification.

For designs that pass simulation, timing quality is compared with golden references (Table~\ref{tab:timing_gap}). The distribution is bimodal rather than uniformly degraded: 6 of 11 tasks show timing ratios within 1.2$\times$ of golden, indicating that LLMs produce timing-adequate designs for most tractable tasks. The timing gap is concentrated in tasks requiring efficient fixed-point division (VWAP: 13.7$\times$, RSI: 3.6$\times$), where LLM-generated designs use naive iterative algorithms in place of optimized shift-based or pipelined implementations. The gap is pattern-specific, not systematic. Figure~\ref{fig:casestudies} (top) provides a VWAP case study, where a single operator choice (\texttt{/} on 48-bit operands) accounts for most of the observed timing collapse.

\subsection{Deployable-but-wrong: bitstream-ready, functionally incorrect HDL}
\label{sec:deployable}

Across the 594 generation attempts, 206 (34.7\%) produce HDL that synthesizes and routes but fails simulation; a further 21 pass simulation but fail to route, and 212 pass both. Among sim-passing designs, 96\% route successfully, so structural validity is nearly free once correctness is achieved---the hazard runs in the other direction. The deployable-but-wrong population concentrates on the trading-indicator (L2) tasks (Table~\ref{tab:deployable}): for RSI, of 10 designs that synthesized and routed, only 1 computes correctly.

The 9 routed-but-wrong RSI designs report plausible post-P\&R metrics (Fmax 3--106\,MHz, 3K--12K LUTs) with no synthesis warnings; in simulation, several always output $\text{RSI}=100$ (passing tests in the $[70, 100]$ range), others use the wrong fixed-point scale (25600 where $[0, 7680]$ is expected, a Q15.8-vs-Q8.8 confusion), and one never asserts \texttt{valid\_out}. An engineer inspecting the post-P\&R report alone would see no toolchain-level signal that these designs are broken. In a financial deployment, such a module could execute trades on bad alpha; an LLM that produces deployable-but-wrong designs is therefore more dangerous than one that fails at synthesis. The evaluation pipeline flags sim-failing designs in its per-design reports so this population is visible without cross-referencing tables.

\subsection{Experiment 2: System configuration}
A fixed 6-stage HFT pipeline~\citep{yoo2023lighttrader,boutros2017hft} (\texttt{UDP\_RX $\to$ FAST\_DEC $\to$ ORDER\_BOOK $\to$ STRATEGY $\to$ RISK\_CHK $\to$ MATCH}) provides 432 configuration combinations across per-stage variants. The LLM does not write code; it selects configurations to minimize end-to-end latency ($= \sum_i \text{depth}_i / \min_i(\text{Fmax}_i)$) under a 5000-LUT budget. Over 24 rounds with per-stage performance feedback, it iteratively refines selections. FPGA results are cached; identical configurations are never re-evaluated. Scale: 6 models $\times$ 5 seeds $\times$ 24 rounds = 720 rounds.

\begin{table}[t]
\begin{minipage}[c]{0.55\textwidth}
\centering
\caption{System DSE results (5 seeds, 24 rounds each). Baselines: Random, Simulated Annealing (SA), Bayesian Optimization (BO/TPE), run with the untuned library-default implementations of Appendix~\ref{app:budget} (values match its $B{=}24$ column). ``Seeds 5/5'' = fraction of seeds reaching the optimal 107.5\,ns.}
\label{tab:exp2_summary}
\footnotesize
\setlength{\tabcolsep}{2.5pt}
\resizebox{\linewidth}{!}{%
\begin{tabular}{lrrrr}
\toprule
Method & Mean best & Std & First opt. & Seeds 5/5 \\
\midrule
\multicolumn{5}{l}{\emph{Baselines}} \\
Random & 128.7 & 11.2 & --- & 0/5 \\
SA & 108.5 & 2.2 & \textbf{R7.5} & 4/5 \\
BO (TPE) & 117.1 & 11.1 & R12.0 & 2/5 \\
\midrule
\multicolumn{5}{l}{\emph{LLMs}} \\
Claude Sonnet 4.6 & \textbf{107.5} & \textbf{0.0} & R11.2 & \textbf{5/5} \\
GPT-5.4 & \textbf{107.5} & \textbf{0.0} & R15.8 & \textbf{5/5} \\
MiniMax M2.7 & 108.5 & 2.2 & R16.8 & 4/5 \\
DeepSeek V3.2 & 116.1 & 11.8 & R12.3 & 3/5 \\
Gemini 3.1 Flash Lite & 121.4 & 10.6 & R17.0 & 1/5 \\
Mistral Large & 129.4 & 25.6 & R23.0 & 1/5 \\
\bottomrule
\end{tabular}}
\end{minipage}\hfill
\begin{minipage}[c]{0.42\textwidth}
\centering
\includegraphics[width=\linewidth]{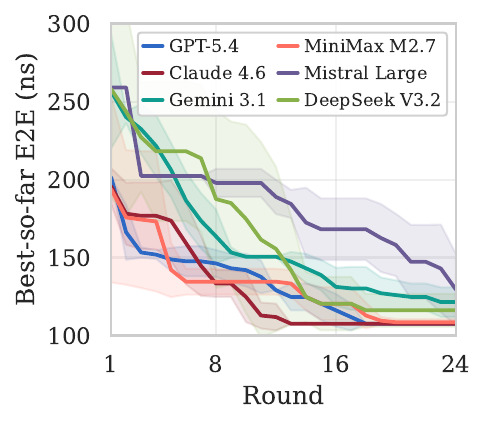}
\captionof{figure}{System DSE convergence over 24 rounds. Mean best-so-far E2E latency across 5 seeds; shading: $\pm$1 std. All models converge toward 107.5\,ns at different rates: Claude by Round~11, Mistral by Round~23.}
\label{fig:dse_convergence}
\end{minipage}
\end{table}

Table~\ref{tab:exp2_summary} and Figure~\ref{fig:dse_convergence} compare LLMs against three baselines: random search, simulated annealing (SA), and Bayesian optimization (BO with TPE sampler). The top LLMs (Claude, GPT-5.4) reach the optimal 107.5\,ns in all 5 seeds, a reliability no baseline achieves at this budget: Random reaches it in 0/5 seeds, SA in 4/5, BO in 2/5. The advantage is not speed---SA, in the seeds where it succeeds, reaches the optimum by Round~7.5 on average, earlier than any LLM---but convergence reliability: only the top LLMs reach the optimum on every seed.

The optimal configuration requires two non-obvious choices: (1) combinational (not pipelined) FAST decoder, saving one pipeline cycle, and (2) DEPTH=4 sequential (not parallel) order book, achieving 46.5\,MHz Fmax despite smaller capacity. Both choices are counterintuitive; the system-level interaction between pipeline depth and global Fmax creates a non-monotonic tradeoff. LLMs learn this through feedback; SA discovers it through local neighborhood exploration; BO remains unreliable at this budget because 24 rounds provide limited data for surrogate model construction in a 432-configuration space. The comparison is budget-conditional: in the sweep of Appendix~\ref{app:budget}, SA reaches 5/5 reliability at 48 rounds and BO at 96, while at the 24-round budget typical when each round costs a full P\&R pass, the two top LLMs are the only 5/5 methods. Figure~\ref{fig:casestudies} (bottom) further contrasts two representative DSE trajectories, showing the difference between directed exploration and local oscillation.

\subsection{Experiment 3: Specification adaptation}
Each LLM receives an existing module's Verilog code, current timing metrics, and a specification change description. Over 3 rounds with feedback, it modifies the code. Fifteen changes span five categories: protocol (4), strategy (4), risk (3), market structure (2), and performance (2). Scale: 6 models $\times$ 15 changes $\times$ 3 seeds = 270 evaluations. Five of the six models returned non-empty responses on all 45 attempts; MiniMax M2.7 returned valid outputs on 17 of 45, with the remaining 28 being service-level failures (the API returned no content), concentrated in the strategy (12/12 missing) and market structure (6/6 missing) categories that require generating longer code modifications.

\begin{table}[t]
\centering
\caption{Top: adaptation success (\%) by category across 15 spec changes, counting passes over \emph{returned} (non-empty) outputs; ``$-$'': no valid output returned. Bottom: for MiniMax, the only model with empty responses, the capability view (pass/returned) vs.\ the deployability view (pass/attempted, 45 attempts).}
\label{tab:exp3_summary}
\begin{tabular}{lrrrrr}
\toprule
Model & Proto. & Strat. & Risk & Mkt.St. & Perf. \\
\midrule
GPT-5.4 & \textbf{67} & 0 & 33 & 50 & 17 \\
Claude Sonnet 4.6 & 50 & \textbf{42} & 33 & 50 & 17 \\
Gemini 3.1 Flash Lite & 50 & 0 & 33 & 0 & 0 \\
MiniMax M2.7 & \textbf{83} & $-$ & 33 & $-$ & 20 \\
DeepSeek V3.2 & 17 & 0 & \textbf{44} & 17 & 0 \\
Mistral Large & 17 & 0 & 33 & 17 & 0 \\
\midrule
MiniMax capability (pass/returned) & 5/6 & 0/0 & 2/6 & 0/0 & 1/5 \\
MiniMax deployability (pass/attempted) & 5/12 & 0/12 & 2/9 & 0/6 & 1/6 \\
\bottomrule
\end{tabular}
\end{table}

The five categories form a clear difficulty gradient (Table~\ref{tab:exp3_summary}). Protocol changes (field counts, bit widths, CRC integration) involve mechanical transformations and are most tractable (GPT: 67\%). Market structure and risk changes require modifying data structure or guard logic and reach 33--50\%. Strategy changes (composing new computational stages such as MACD or Greeks-based hedging) are solved only by Claude (42\%). Performance optimization (pipeline insertion for Fmax) remains near-zero (0--20\%); the one partial success is binomial step doubling, a parameter change rather than a microarchitectural transformation.

\subsection{Cross-experiment analysis}

\begin{table}[t]
\centering
\caption{Cross-experiment comparison. Spearman correlations between generation and DSE are reported along both DSE quality axes: convergence speed (first-optimum round) $\rho=-0.55$ and convergence reliability (seeds reaching 107.5\,ns) $\rho=+0.54$, both $p=0.27$ (exact permutation, $n=6$). Gen vs.\ Adapt: $\rho=0.66$. Adaptation Sim\% counts passes over returned (non-empty) outputs; see Table~\ref{tab:exp3_summary}.}
\label{tab:cross_exp}
\begin{tabular}{lcccccc}
\toprule
 & \multicolumn{2}{c}{Generation} & \multicolumn{2}{c}{DSE} & \multicolumn{2}{c}{Adaptation} \\
\cmidrule(lr){2-3} \cmidrule(lr){4-5} \cmidrule(lr){6-7}
Model & Sim\% & Rk & First opt. & Rk & Sim\% & Rk \\
\midrule
Claude Sonnet 4.6      & \textbf{61} & 1 & \textbf{R11.2} & 1 & \textbf{37} & 1 \\
GPT-5.4         & \textbf{61} & 1 & R15.8          & 3 & 27          & 2 \\
Gemini 3.1 Flash Lite      & 46          & 3 & R17.0          & 5 & 13          & 4 \\
DeepSeek V3.2   & 25          & 4 & R12.3          & 2 & 10          & 5 \\
Mistral Large   & 23          & 5 & R23.0          & 6 & 7           & 6 \\
MiniMax M2.7    & 19          & 6 & R16.8          & 4 & 20          & 3 \\
\bottomrule
\end{tabular}
\end{table}

Because Table~\ref{tab:exp2_summary} emphasizes convergence \emph{reliability} while ranking models by convergence \emph{speed} is equally natural, the generation-vs.-DSE correlation is reported along both axes. The two agree in capability terms: convergence speed gives $\rho=-0.55$ (better generator, faster DSE) and seed reliability gives $\rho=+0.54$ (better generator, more reliable DSE), both $p=0.27$ under exact permutation testing at $n=6$. A magnitude of $|\rho|\approx0.55$ is a moderate correlation by standard convention, so the two rankings overlap moderately rather than being independent; at $n=6$ the sign cannot be established with confidence, and a larger model sample would be needed to sharpen the estimate. The informative signal is in the per-model exceptions: DeepSeek ranks fourth in generation but second in DSE convergence speed, and MiniMax ranks last in generation but third in seed reliability. Generation and adaptation rankings show moderate agreement ($\rho = 0.66$, $p=0.16$), consistent with both requiring Verilog production. The practical observation remains: the best code generator (Claude) is not the fastest architecture optimizer, and the worst code generator (MiniMax) still reaches near-optimal DSE results.

\begin{figure}[t]
\begin{tcolorbox}[colback=csviolet!5, colframe=csviolet, title={\textbf{Case Study 1: VWAP -- Why a Single Operator Choice Creates 13.67$\times$ Timing Degradation}}, fonttitle=\small, coltitle=white, boxsep=1pt, left=3pt, right=3pt, top=2pt, bottom=2pt]
\small
VWAP = $\sum$(\text{price} $\times$ \text{volume}) $/$ $\sum$(\text{volume}), with 48-bit accumulators. Of 18 LLM-generated designs (6 models $\times$ 3 trials), 2 pass simulation; 14 route with reg2reg critical paths of 140--312\,ns, all from one root issue: the Verilog \texttt{/} operator for 48-bit combinational division, which Yosys maps to a subtract-compare chain with $>$100 logic levels.

\begin{minipage}[t]{0.48\textwidth}
\begin{tcolorbox}[colback=csgold!15, colframe=csgold!80!black, title={\scriptsize \textbf{Golden Reference} iterative FSM, 5.3\,ns, 46.5\,MHz}, fonttitle=\scriptsize, boxsep=1pt, left=2pt, right=2pt, top=1pt, bottom=1pt]
\begin{lstlisting}[language=verilog, aboveskip=1pt, belowskip=1pt]
reg [47:0] cumulative_pv, cumulative_vol;
// Accumulate on each valid trade:
cumulative_pv  <= cumulative_pv + pv_product;
cumulative_vol <= cumulative_vol + vol_ext;
// Division via restoring FSM:
//   IDLE -> CALC (48 cycles) -> DONE
//   one quotient bit/cycle, no comb /
\end{lstlisting}
\end{tcolorbox}
\end{minipage}
\hfill
\begin{minipage}[t]{0.48\textwidth}
\begin{tcolorbox}[colback=cspeach!25, colframe=cspeach!80!black, title={\scriptsize \textbf{GPT-5.4} combo division, 307.7\,ns, 3.0\,MHz}, fonttitle=\scriptsize, boxsep=1pt, left=2pt, right=2pt, top=1pt, bottom=1pt]
\begin{lstlisting}[language=verilog, aboveskip=1pt, belowskip=1pt]
// 48-bit accumulators (same as golden)
reg [47:0] cum_pv, next_cum_vol;
// But division is combinational:
assign vwap_calc = (next_cum_vol != 0)
    ? (dividend / next_cum_vol) : 48'd0;
// Yosys: 48-bit / -> >100 logic levels
//   Fmax collapses from 46 to 3 MHz
\end{lstlisting}
\end{tcolorbox}
\end{minipage}

{\scriptsize The \texttt{/} produces the same collapse in 16/18 designs (Claude t=0 passes simulation, 5.07\,ns, because its accumulator logic differs); only iterative division avoids it.}
\end{tcolorbox}

\vspace{-4pt}

\begin{tcolorbox}[colback=cslilac!10, colframe=cslilac!80!black, title={\textbf{Case Study 2: DSE -- Directed Exploration vs.\ Stuck Oscillation (seed 0, rounds 1--15)}}, fonttitle=\small, boxsep=1pt, left=3pt, right=3pt, top=2pt, bottom=2pt]
\small
\begin{minipage}[t]{0.48\textwidth}
\begin{tcolorbox}[colback=csgold!10, colframe=csgold!80!black, title={\scriptsize \textbf{Claude Sonnet 4.6}: 173\,ns $\to$ 107.5\,ns in 13 rounds}, fonttitle=\scriptsize, boxsep=1pt, left=2pt, right=2pt, top=1pt, bottom=1pt]
{\scriptsize\setlength{\tabcolsep}{2pt}
\begin{tabular}{rll}
R1 & wide / pipe / \textbf{par\_8} & $\to$ 173\,ns\\
R2 & \textbf{medium} / pipe / seq\_8 & $\to$ 270\,ns\\
R3 & \textbf{narrow} / pipe / par\_4 & $\to$ 292\,ns\\
R6 & \textbf{wide} / pipe / par\_8 & $\to$ 202\,ns\\
R7 & wide / pipe / \textbf{par\_4} & $\to$ 135\,ns\\
R8 & wide / pipe / \textbf{seq\_4} & $\to$ \textbf{129}\,ns\\
R13 & wide / \textbf{comb} / seq\_4 & $\to$ \textbf{107.5}\,ns\,$\checkmark$\\
\end{tabular}

\vspace{1pt}
\color{gray}{R1--5 explore all bus widths; R6 returns to wide; R7--8 cut S2 depth to seq\_4; R13 drops the S1 pipeline $\to$ optimum.}
}
\end{tcolorbox}
\end{minipage}
\hfill
\begin{minipage}[t]{0.48\textwidth}
\begin{tcolorbox}[colback=cspeach!15, colframe=cspeach!80!black, title={\scriptsize \textbf{Mistral Large}: stuck at 202\,ns for 20 rounds}, fonttitle=\scriptsize, boxsep=1pt, left=2pt, right=2pt, top=1pt, bottom=1pt]
{\scriptsize\setlength{\tabcolsep}{2pt}
\begin{tabular}{rll}
R1 & medium / pipe / \textbf{par\_8} & $\to$ 259\,ns\\
R2 & medium / pipe / \textbf{par\_16} & $\to$ 306\,ns\\
R3 & medium / pipe / \textbf{par\_4} & $\to$ 202\,ns\\
R4--10 & medium / pipe / par\_8$\leftrightarrow$4 & $\to$ 259$\leftrightarrow$202\,ns\\
\end{tabular}

\vspace{1pt}
\color{gray}{S0 never leaves medium, S1 never tests comb; only S2 oscillates between par\_4 and par\_8. The optimum requires changing all three stages at once.}
}
\end{tcolorbox}
\end{minipage}
\end{tcolorbox}
\vspace{-4pt}
\caption{Case studies. (Top) VWAP: the Verilog \texttt{/} operator on 48-bit operands collapses Fmax from 46.5 to 3.0\,MHz; only iterative division avoids this. (Bottom) DSE: Claude explores broadly (R1--R5), identifies the bottleneck (R6--R8), then optimizes globally (R13). Mistral oscillates on a single knob for 20 rounds without questioning its other choices.}
\label{fig:casestudies}
\end{figure}

\section{Discussion}
\label{sec:discussion}

\textbf{Three capabilities and training data as difficulty predictor.}
The three capabilities draw on different strengths---syntactic pattern matching for code generation, quantitative optimization from numerical feedback for system configuration, and program comprehension for adaptation---consistent with the moderate, rather than perfect, overlap of cross-experiment rankings. Timing degradation concentrates in tasks requiring efficient fixed-point division (VWAP: 13.7$\times$; RSI: 3.6$\times$), where open-source Verilog training data contains only naive implementations. This pattern recurs across the benchmark: tasks with well-represented canonical implementations are tractable, while tasks whose efficient implementations are rarely published in open-source corpora (pipelined division, parallel book traversal, transcendental approximation) tend to remain unsolved by current models. The DSE results further suggest that LLMs are effective not because they predict timing from first principles, but because they provide a useful search prior that is progressively refined by structured feedback. The moderate overlap of capabilities suggests that production LLM-assisted hardware design tools should not rely on a single model: an architecture-search model could select configurations, a code-generation model could implement them, and a domain expert could handle adaptation tasks where all models fail.

\textbf{Service availability as a separate axis.}
In a trading deployment, a model that returns no response is operationally equivalent to one that returns a wrong response---the engineer is blocked either way. MiniMax's 83\% protocol-adaptation capability drops to 42\% deployability once service-level failures are counted (Table~\ref{tab:exp3_summary} reports both); for production LLM use in latency-critical domains, service availability and model capability should be tracked as separate axes.

\textbf{Limitations.}
The main results use Lattice ECP5 with open-source tools for reproducibility, while deployed HFT cards use Xilinx Virtex UltraScale+ parts; a Vivado pilot on a curated 13-design subset finds high Fmax rank agreement between the flows ($\rho=0.982$, Appendix~\ref{app:vivado}), but broader replication across designs, parts, and toolchains is open work. Sim\% reflects RTL-level testbench simulation---no gate-level simulation or FPGA-in-the-loop testing---and testbench corner-case coverage is future work. The Exp.~2 latency model is purely structural ($\text{E2E} = \sum_i \text{depth}_i / \min_i \text{Fmax}_i$): no queue fill, stall propagation, or data-dependent timing, with the strategy stage idealized as 1-cycle---most consequential at the \texttt{ORDER\_BOOK}$\to$\texttt{STRATEGY} handoff, where production systems buffer bursty updates; a backpressure-aware extension with per-stage FIFO depths as DSE dimensions is a direct next iteration. The 432-configuration space is modest by DSE standards; how the LLM-vs-classical sample-efficiency gap scales with dimensionality is unquantified. Six general-purpose LLMs are tested; HDL-specialized models (VeriGen~\citep{thakur2023verigen}, RTLCoder~\citep{liu2024rtlcoder}) are superseded by current frontier models~\citep{pinckney2025cvdp}. Cross-experiment correlation ($n=6$) has limited statistical power. Golden references are verified through automated testbenches and FPGA P\&R; option pricing is validated against analytical solutions (BS error: 0.02\%).

\section{Related Work}
\label{sec:related}

\textbf{HDL generation benchmarks.}
VerilogEval~\citep{liu2023verilogeval} (156 tasks from HDLBits tutorials) and RTLLM~\citep{lu2024rtllm} (50 specification-to-RTL tasks) measure functional correctness on general-purpose digital logic. ResBench~\citep{guo2025resbench} is the closest prior work, evaluating LUT utilization across 56 FPGA tasks including a financial computing category, but measures resource efficiency rather than timing. Table~\ref{tab:benchmarks} summarizes the differences: FinHardBench adds post-P\&R timing at domain-specific critical paths, parameterized knobs enabling system-level DSE, and specification adaptation.

\textbf{LLM-based alpha factor mining.}
The L2 (trading indicators) and L5 (option/risk) tasks implement financial alpha computations, and a parallel literature studies LLMs at the software level: mining formulaic alpha factors from historical time series. AlphaForge~\citep{shi2025alphaforge} mines and dynamically combines formulaic factors; AlphaAgent~\citep{tang2025alphaagent} adds LLM-driven multi-agent exploration to counteract alpha decay; AlphaBench~\citep{luo2026alphabench}, concurrent work released after the submission deadline, benchmarks LLM-based factor generation, evaluation, and searching. FinHardBench addresses the downstream problem: given a financial computation---a mined factor or a canonical indicator such as RSI or VWAP---can an LLM implement it as latency-aware FPGA hardware? The two are complementary: a discovered factor still needs correct, fast HDL to deploy at HFT timescales, and current LLMs make errors on the hardware-implementation problem that software-level factor-mining benchmarks do not surface.

\textbf{LLM-based hardware optimization.}
VeriPPA~\citep{thorat2023verippa} uses multi-round LLM refinement to optimize power-performance-area of generated Verilog, evaluated on Synopsys Design Compiler with a 7\,nm PDK. FinHardBench adopts a similar feedback loop structure for Exp.~2 (system DSE) and Exp.~3 (adaptation), but targets FPGA timing rather than ASIC PPA and operates at the system level rather than single-module level. Mercury~\citep{du2024mercury} benchmarks runtime efficiency of LLM-generated software code, introducing the Beyond@$k$ metric. The Fmax ratio and timing ratio metrics in FinHardBench serve an analogous role for hardware.

\textbf{Financial FPGA systems.}
\citet{boutros2017hft} implement a complete HFT system using HLS, achieving $<$870\,ns round-trip. \citet{lockwood2012fpga} provide an FPGA IP library supporting FIX, OUCH, and BATS BOE protocols. \citet{oliveira2023fast} open-source a FAST decoder IP core. \citet{mahony2024fpga} survey 99 studies on FPGA-accelerated option pricing, reporting 270--5400$\times$ speedups over CPUs. \citet{yoo2023lighttrader} present LightTrader, integrating DNN accelerators into an HFT pipeline; the FinHardBench pipeline topology follows their architecture. All prior financial FPGA work is hand-designed.

\textbf{LLM for design space exploration.}
LLM-guided optimization has been applied to neural architecture search~\citep{fang2026llmnas} and chip design assistants~\citep{liu2023chipnemo}, but not to hardware microarchitectural configuration with FPGA timing feedback. Exp.~2 of FinHardBench provides the first such evaluation, with baseline comparisons showing that top LLMs achieve superior convergence reliability.

\section{Conclusion}
\label{sec:conclusion}

FinHardBench evaluates LLMs on three stages of the financial FPGA iteration cycle: module generation, system configuration, and specification adaptation. The system DSE experiment reveals that every tested model reaches the same optimal latency in at least one seed within 24 rounds, but convergence speed varies by 2$\times$ (Claude: Round~11, Mistral: Round~23). The top LLMs achieve 5/5 seed reliability, exceeding random search (0/5), SA (4/5), and BO (2/5) at the same budget. Across the six models, generation and DSE rankings overlap moderately under both DSE metrics tested ($|\rho| \approx 0.55$ for convergence speed and seed reliability), with informative per-model exceptions: DeepSeek ranks 4th in generation but 2nd in DSE convergence speed, and MiniMax ranks 6th in generation but 3rd in seed reliability. Strategy adaptation and timing optimization remain unsolved for most models in this evaluation. FinHardBench is released as open-source to support research on each dimension independently.

\section*{Acknowledgments}

This work was supported in part by the National Science Foundation under Grants 2340949, 2419880, and 2613797, and by the NYUAD Center for Cyber Security (CCS) through the NYUAD Research Institute Award G1104 funded by Tamkeen. The authors acknowledge the use of High Performance Computing resources at New York University Abu Dhabi and Kansas State University.

\section*{Reproducibility Statement}

All benchmark tasks (33 specifications, golden references, testbenches, critical I/O definitions, and knob configurations), evaluation scripts, system pipeline variants, and experiment drivers are available at \url{https://github.com/owenfucell/FinHardBench}. The evaluation toolchain consists of open-source components: Icarus Verilog 12.0 for simulation, Yosys 0.57 for synthesis, and NextPNR for Lattice ECP5-85K P\&R. Experiment results are stored in SQLite databases included in the repository. LLM evaluations use the OpenRouter API with model identifiers and temperature (0.7) specified in \texttt{models.json}. Random search, simulated annealing, and Bayesian optimization baselines are implemented in \texttt{run\_dse.py} with fixed seeds for reproducibility.

\bibliography{colm2026_conference_new}
\bibliographystyle{colm2026_conference}

\appendix

\section{Cross-toolchain validation on Vivado / Kintex UltraScale+}
\label{app:vivado}

To test whether the timing trends measured on the open ECP5 flow transfer to a commercial flow, 13 designs (golden references, key LLM failures, and DSE per-stage variants) were re-synthesized with Vivado 2025.2 on the Kintex UltraScale+ part \texttt{xcku5p-ffvb676-2-e}, using the same Verilog sources evaluated on ECP5+NextPNR. The cross-toolchain script and results are in the repository (\texttt{system/run\_vivado.py}, \texttt{system/vivado\_results.csv}, \texttt{system/cross\_toolchain\_report.md}). Table~\ref{tab:vivado} reports the 11 successful pairings.

\begin{table}[h]
\centering
\caption{ECP5 (Yosys+NextPNR) vs.\ Vivado (Kintex UltraScale+) Fmax for the cross-toolchain subset. Spearman rank correlation across paired designs: $\rho=0.982$.}
\label{tab:vivado}
\small
\begin{tabular}{lrrr}
\toprule
Design & ECP5 Fmax (MHz) & Vivado Fmax (MHz) & Ratio \\
\midrule
VWAP golden             & 215.94 & 981.35  & 4.54 \\
VWAP GPT-5.4 t=2 (\texttt{/}) & 2.95 & 22.11 & 7.49 \\
SMA golden              & 131.77 & 482.63  & 3.66 \\
EMA golden              & 76.68  & 383.14  & 5.00 \\
RSI golden              & 10.39  & 170.39  & 16.40 \\
RSI Claude-4.6 t=1      & 3.88   & 31.61   & 8.15 \\
ORDER\_BOOK\_SIDE golden & 33.32 & 217.06  & 6.51 \\
S2 seq\_4 (DSE-optimal) & 46.52  & 231.75  & 4.98 \\
S2 par\_8               & 34.73  & 215.70  & 6.21 \\
S1 comb (DSE-optimal)   & 713.78 & 1222.49 & 1.71 \\
S1 pipelined            & 640.61 & 1447.18 & 2.26 \\
\bottomrule
\end{tabular}
\end{table}

Absolute Fmax scales by 1.7--16.4$\times$ between the two flows, but the rank ordering of designs is nearly preserved ($\rho=0.982$): designs that are slow on ECP5 are slow on UltraScale+, and the LLM-induced timing collapses (VWAP with combinational \texttt{/}: 73$\times$ gap on ECP5, 44$\times$ on Vivado) reproduce on the commercial flow. One reversal appears at stage level: the S1 combinational and pipelined variants swap rank between toolchains; this does not change the system-level optimum. Two designs failed to pair: \texttt{BS\_CALL\_PRICE} golden (Vivado synthesis non-convergence) and its Gemini variant (timing-report timeout). Deployed HFT cards use the Virtex UltraScale+ family (per AMD Alveo UL3422/UL3524 material and the STAC-T0 benchmark); firm-specific configurations are not public, so this 14-design pilot is evidence of transfer, not proof.

\section{DSE baseline configuration and budget sensitivity}
\label{app:budget}

\textbf{Baseline hyperparameters.} All three baselines (implemented in \texttt{system/baselines.py}) operate on the same per-stage P\&R cache as the LLM experiments and use documented library defaults with no tuning, mirroring the out-of-the-box use of the LLMs. \emph{Random}: \texttt{numpy.random.default\_rng(seed)}, uniform without replacement over the 432-point space. \emph{Simulated annealing}: single-knob-flip neighborhood, exponential cooling from $T_0=2.0$ to $T_{\min}=0.01$ (cooling-rate defaults from SciPy's \texttt{dual\_annealing} reference), Metropolis acceptance, uniform random start. \emph{BO (TPE)}: Optuna's \texttt{TPESampler} with defaults (\texttt{n\_startup\_trials}=10, \texttt{n\_ei\_candidates}=24, \texttt{multivariate}=False), each per-stage knob a categorical dimension.

\textbf{Budget sensitivity.} The three baselines were run at budgets $B \in \{12, 24, 48, 96\}$ (5 seeds each), and the two top LLMs were extended from 24 to 48 rounds; the first 24 rounds of each LLM run are bit-identical to Table~\ref{tab:exp2_summary}. Table~\ref{tab:budget} reports mean best-so-far latency and the number of seeds reaching the 107.5\,ns optimum.

\begin{table}[h]
\centering
\caption{Budget sweep: mean best E2E latency (ns) / seeds reaching 107.5\,ns, at four round budgets. Bold marks the first budget at which a method reaches the optimum on all 5 seeds.}
\label{tab:budget}
\small
\begin{tabular}{lcccc}
\toprule
Method & $B$=12 & $B$=24 & $B$=48 & $B$=96 \\
\midrule
Random      & 151.3 / 0 & 128.7 / 0 & 116.1 / 3 & 111.8 / 4 \\
SA          & 119.9 / 4 & 108.5 / 4 & \textbf{107.5 / 5} & 107.5 / 5 \\
BO (TPE)    & 130.2 / 1 & 117.1 / 2 & 117.1 / 2 & \textbf{107.5 / 5} \\
Claude 4.6  & 111.8 / 4 & \textbf{107.5 / 5} & 107.5 / 5 & 107.5 / 5 \\
GPT-5.4     & 129.0 / 0 & \textbf{107.5 / 5} & 107.5 / 5 & 107.5 / 5 \\
\bottomrule
\end{tabular}
\end{table}

The reading is budget-conditional: at $B$=24---roughly what a practitioner runs when each round costs a full P\&R pass---the two LLMs are the only methods at 5/5 reliability; SA catches up at $B$=48 and BO at $B$=96, while Random does not reach 5/5 even at $B$=96. The 24-round budget was set by per-round toolchain wall-clock cost, not chosen to favor the LLMs.

\section{L5 (Options \& Risk) task specifications and failure analysis}
\label{app:l5}

Table~\ref{tab:l5_tasks} covers all ten L5 tasks: the computation, the critical-I/O path used for timing, the testbench check, and the simulation pass rate across the 18 generation attempts (6 models $\times$ 3 trials). The six guard/P\&L tasks use exact-match fixed-point testbenches and are largely tractable; failures concentrate in the four option pricing tasks, whose tolerance-based testbenches check against analytical references.

\begin{table}[h]
\centering
\caption{All ten L5 tasks. Critical-I/O paths are per-task (\texttt{critical\_io.json}); for the multi-output \texttt{GREEKS\_BUNDLE}, the path is taken to the last-asserted output among delta/gamma/theta. Pricing testbenches check within a stated tolerance ($\pm5\%$ for \texttt{BS\_CALL\_PRICE}) against analytical references; guard/P\&L testbenches are exact-match in fixed point.}
\label{tab:l5_tasks}
\small
\setlength{\tabcolsep}{3.5pt}
\begin{tabular}{llllr}
\toprule
Task & Computation & Critical path & Check & Sim pass \\
\midrule
POS\_LIMIT & Position limit checker & \texttt{order\_qty}$\to$\texttt{allow} & exact & 15/18 \\
PRICE\_GUARD & Price anomaly detector & \texttt{current\_price}$\to$\texttt{alert} & exact & 18/18 \\
RATE\_LIMIT & Token-bucket rate limiter & \texttt{request}$\to$\texttt{allow} & exact & 10/18 \\
PNL\_CALC & Mark-to-market P\&L & \texttt{valid\_in}$\to$\texttt{valid\_out} & exact & 9/18 \\
MULTI\_LEG\_PNL & Multi-leg position P\&L & \texttt{valid\_in}$\to$\texttt{valid\_out} & exact & 8/18 \\
COMBO\_PAYOFF & Combo expiry payoff & \texttt{valid\_in}$\to$\texttt{valid\_out} & exact & 10/18 \\
BS\_CALL\_PRICE & Black-Scholes call price & \texttt{valid\_in}$\to$\texttt{valid\_out} & tolerance & 0/18 \\
OPTION\_DELTA & Black-Scholes delta & \texttt{valid\_in}$\to$\texttt{valid\_out} & tolerance & 2/18 \\
GREEKS\_BUNDLE & Delta + gamma + theta & \texttt{valid\_in}$\to$last output & tolerance & 1/18 \\
BINOM\_PRICER & Binomial tree pricer & \texttt{valid\_in}$\to$\texttt{valid\_out} & tolerance & 2/18 \\
\bottomrule
\end{tabular}
\end{table}

L5 constitutes 10 of the 33 tasks. Each L5 task ships with the module name, parameter list (e.g., \texttt{WIDTH=32}, \texttt{FRAC=16} for Q16.16 fixed-point), complete port declarations, and a functional specification giving the closed-form formula and the algorithmic decomposition of each transcendental component: $N(x)$ via the Abramowitz--Stegun polynomial approximation; $\ln(x)$ via $2\,\mathrm{atanh}((x-1)/(x+1))$ with polynomial expansion; $\sqrt{x}$ via Newton--Raphson with 5 iterations; $e^{-x}$ via Taylor series with repeated-squaring range reduction. Numerical tolerance is $\pm5\%$ against the analytical reference, stated in the specification text---generous relative to the precision floor of Q16.16. The critical-I/O path is \texttt{valid\_in}$\to$\texttt{valid\_out} for single-output tasks (\texttt{BS\_CALL\_PRICE}, \texttt{OPTION\_DELTA}, \texttt{BINOM\_PRICER}); for the multi-output \texttt{GREEKS\_BUNDLE}, the path is taken to the last-asserted output among delta/gamma/theta, giving worst-case end-to-end latency.

\textbf{Failure modes on \texttt{BS\_CALL\_PRICE}.} Across 18 attempts (6 models $\times$ 3 trials): 9 did not produce a complete, syntactically closed module (the golden reference is 9.9\,KB; failing to express a working implementation at that scale is a conciseness-and-planning failure that would translate into disproportionate area in deployment); 5 produced Verilog-level errors (invalid l-values, undeclared variables); 4 produced bitstream-ready designs that route cleanly but compute incorrect values (Gemini Flash Lite $\times$3, DeepSeek V3.2 $\times$1).

The 4 routed-but-wrong cases disambiguate the candidate explanations for the 0\% pass rate. The representative case (Gemini Flash Lite, trial 1) systematically substituted the specification-named algorithms with simpler approximations: $N(x)$ (spec: Abramowitz--Stegun) implemented as the logistic sigmoid $1/(1+e^{-1.6x})$, a different function with $\approx$2\% peak error at $x=0$ growing in the tails; $\ln(S/K)$ (spec: $2\,\mathrm{atanh}$ form) implemented as the first-order expansion $(S-K)/K$, valid only for $S \approx K$; $e^{-x}$ (spec: Taylor with range reduction) implemented as a 4-term Taylor series without range reduction, divergent for $x>1$; only $\sqrt{T}$ (Newton--Raphson) was implemented as specified. These are not tolerance failures (a logistic sigmoid is a different function from the Gauss CDF) and not specification failures (the spec names the canonical algorithms explicitly). The 0\% L5 pass rate therefore reflects a genuine capability gap---when the specification names a canonical algorithm, current models substitute a simpler training-data approximation---rather than an overly strict testbench or an under-specified prompt. The other three routed-but-failing cases show the same substitution pattern.

\section{Agentic pilot}
\label{app:agentic}

Two off-the-shelf coding agents (Claude Code 2.1.158 with \texttt{claude-opus-4-7}, default thinking budget; Codex 0.130.0 with \texttt{gpt-5.5}, reasoning effort \texttt{xhigh}) were given access to \texttt{iverilog} and \texttt{vvp} and asked to implement and verify three tasks autonomously. Prompts and artifacts are in the repository.

\begin{table}[h]
\centering
\caption{Agentic pilot vs.\ single-shot generation. Single-shot Sim\% pools all 18 Exp.~1 trials (one LLM call each).}
\label{tab:agentic}
\small
\begin{tabular}{lccc}
\toprule
Task & Single-shot Sim\% & Claude Code & Codex \\
\midrule
SMA  & 78\% (14/18) & pass $\cdot$ 8 calls / 59\,s  & pass $\cdot$ 4 calls / 70\,s \\
VWAP & 11\% (2/18)  & pass $\cdot$ 8 calls / 47\,s  & pass $\cdot$ 5 calls / 66\,s \\
RSI  & 6\% (1/18)   & pass $\cdot$ 9 calls / 171\,s & pass $\cdot$ 7 calls / 130\,s \\
\bottomrule
\end{tabular}
\end{table}

Both agents pass all three tasks, including VWAP and RSI, which fail most consistently in single-shot mode. Two caveats bound the reading. First, the agent backbones are stronger and more recent than the Exp.~1 models, so the pilot conflates iteration with a backbone upgrade; an apples-to-apples agent run with the Exp.~1 backbones is future work. Second, the fair-budget axis is LLM calls per task: 1 in Exp.~1, 4--9 in the pilot (spent iterating against \texttt{iverilog}+\texttt{vvp} output), 24 per seed in Exp.~2, and 3 per seed in Exp.~3. The pilot measures iteration with autonomous tool access, and the agent prompt was not optimized for Fmax, so the observation is restricted to correctness. The result is consistent with iteration---rather than the specific harness---being what lifts hard-task correctness. The submitted experiments retain fixed prompting because agentic configurations make the unit of measurement a model-plus-framework pair; a full agentic evaluation answers the complementary question of which combination performs best and is a natural follow-on iteration of the benchmark.

\end{document}